\documentclass[times, review, 10pt]{elsarticle}
\usepackage{amssymb}
\usepackage{amsmath}
\usepackage{tikz}
\usepackage{amsmath}
\usepackage{pifont}
\usepackage{makecell}
\usepackage{colortbl}
\usepackage{hanging}
\usepackage{multirow}
\usepackage{graphicx}
\usepackage{subfig}

\usepackage{algorithm}
\usepackage{algorithmicx}
\usepackage{algpseudocode}
\usepackage[utf8]{inputenc}
\usepackage{lmodern}

\journal{Pattern Recognition}

\begin{document}

\begin{frontmatter}

%% Title, authors and addresses

%% use the tnoteref command within \title for footnotes;
%% use the tnotetext command for theassociated footnote;
%% use the fnref command within \author or \affiliation for footnotes;
%% use the fntext command for theassociated footnote;
%% use the corref command within \author for corresponding author footnotes;
%% use the cortext command for theassociated footnote;
%% use the ead command for the email address,
%% and the form \ead[url] for the home page:
%% \title{Title\tnoteref{label1}}
%% \tnotetext[label1]{}
%% \author{Name\corref{cor1}\fnref{label2}}
%% \ead{email address}
%% \ead[url]{home page}
%% \fntext[label2]{}
%% \cortext[cor1]{}
%% \affiliation{organization={},
%%             addressline={},
%%             city={},
%%             postcode={},
%%             state={},
%%             country={}}
%% \fntext[label3]{}

\title{SSL-SSAW: Self-Supervised Learning with Sigmoid Self-Attention Weighting for Question-Based Sign Language Translation} %% Article title

%% use optional labels to link authors explicitly to addresses:
%% \author[label1,label2]{}
%% \affiliation[label1]{organization={},
%%             addressline={},
%%             city={},
%%             postcode={},
%%             state={},
%%             country={}}
%%
%% \affiliation[label2]{organization={},
%%             addressline={},
%%             city={},
%%             postcode={},
%%             state={},
%%             country={}}

\author[label1]{Zekang Liu} %% Author name
\author[label1]{Wei Feng}
\author[label1]{Fanhua Shang}
\author[label1]{Lianyu Hu}
\author[label1]{Jichao Feng}
\author[label2]{Liqing Gao} 

%% Author affiliation
\affiliation[label1]{organization={School of Computer Science and Technology, College of Intelligence and Computing, Tianjin University}, 
	addressline={No. 135 Yaguan Road, Jinnan District}, 
	city={Tianjin},
	postcode={300350}, 
	state={Tianjin},
	country={China}}

\affiliation[label2]{organization={School of Computer Science and Technology, Tiangong University}, 
	addressline={No. 399 Binshui West Road, Xiqing District}, 
	city={Tianjin},
	postcode={300387}, 
	state={Tianjin},
	country={China}}

%% Abstract
\begin{abstract}
Sign Language Translation (SLT) bridges the communication gap between deaf people and hearing people, where dialogue provides crucial contextual cues to aid in translation. Building on this foundational concept, this paper proposes Question-based Sign Language Translation (QB-SLT), a novel task that explores the efficient integration of dialogue. Unlike gloss (sign language transcription) annotations, dialogue naturally occurs in communication and is easier to annotate. The key challenge lies in aligning multimodality features while leveraging the context of the question to improve translation. To address this issue, we propose a cross-modality Self-supervised Learning with Sigmoid Self-attention Weighting (SSL-SSAW) fusion method for sign language translation. Specifically, we employ contrastive learning to align multimodality features in QB-SLT, then introduce a Sigmoid Self-attention Weighting (SSAW) module for adaptive feature extraction from question and sign language sequences. Additionally, we leverage available question text through self-supervised learning to enhance representation and translation capabilities. We evaluated our approach on newly constructed CSL-Daily-QA and PHOENIX-2014T-QA datasets, where SSL-SSAW achieved SOTA performance. Notably, easily accessible question assistance can achieve or even surpass the performance of gloss assistance. Furthermore, visualization results demonstrate the effectiveness of incorporating dialogue in improving translation quality.
\end{abstract}

%%Graphical abstract
\begin{graphicalabstract}
\includegraphics[scale=0.5]{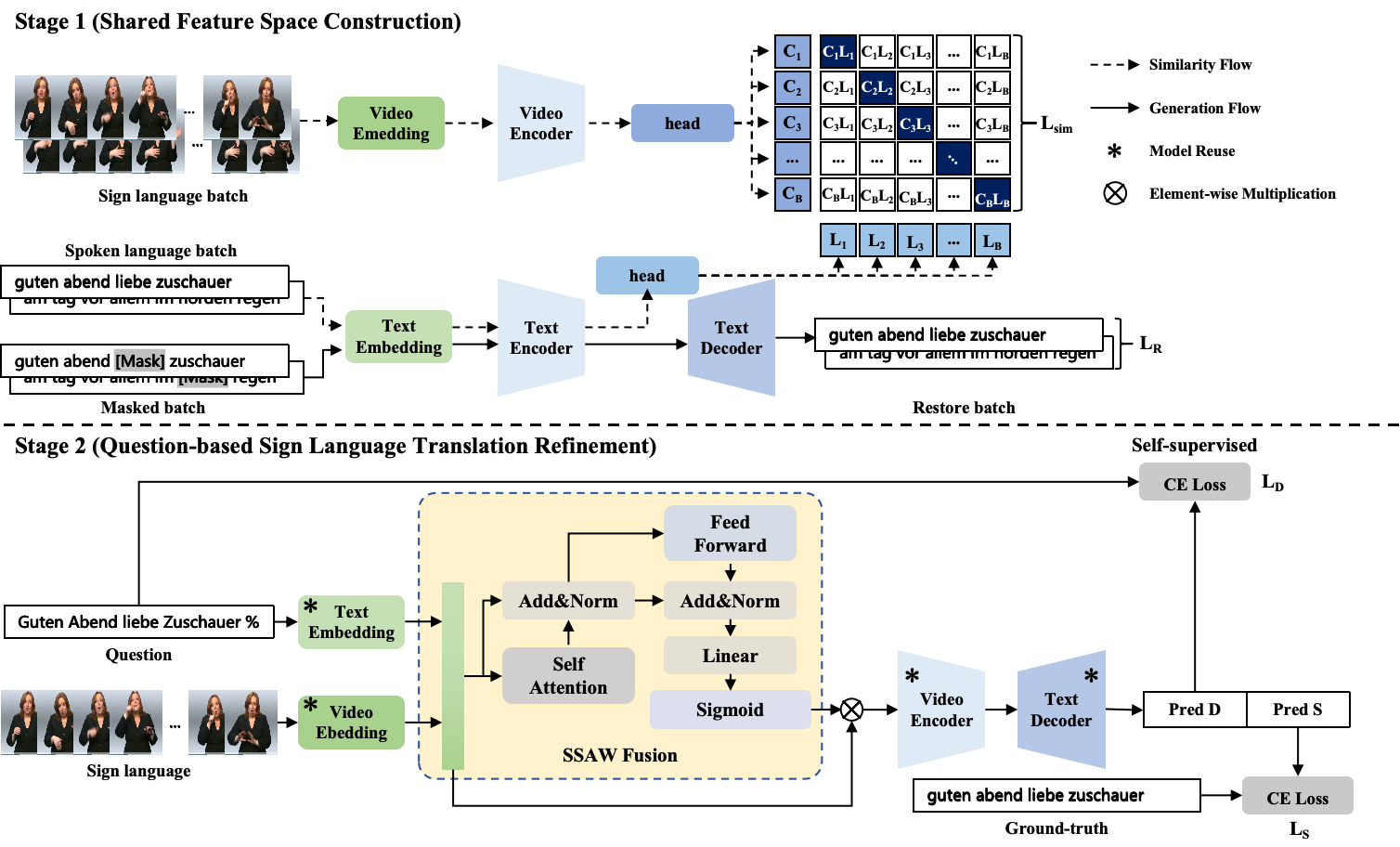}
\end{graphicalabstract}

%%Research highlights
\begin{highlights}
	\item Replacing gloss with low-cost question text exceeds gloss-based translation accuracy.

	\item The proposed SSAW uses question-adaptive key frame selection for improved translation.

	\item We use question for self-supervised learning to enhance context and generalization.
\end{highlights}

%% Keywords
\begin{keyword}
%% keywords here, in the form: keyword \sep keyword

%% PACS codes here, in the form: \PACS code \sep code

%% MSC codes here, in the form: \MSC code \sep code
%% or \MSC[2008] code \sep code (2000 is the default)

Question-based Sign Language Translation  \sep Cross-modality Self-supervised Learning \sep Sigmoid Self-attention Weighting

\end{keyword}

\end{frontmatter}

%% Add \usepackage{lineno} before \begin{document} and uncomment 
%% following line to enable line numbers
%% \linenumbers

%% main text
%%

%% Use \section commands to start a section
\section{Introduction}
Sign language is a natural language used for communication that has its own distinct set of vocabulary and grammar rules. However, unlike spoken language, sign language is a visual language that relies on visual cues such as the shape, position, and motion trajectory of hands and arms, as well as emotional expressions conveyed through facial expressions and motion extent. Sign language translation involves converting sign language videos into sentences, following the order of spoken language, thereby bridging the communication gap between deaf people and hearing people. 

Gloss \footnote{Gloss is an annotation method that uses written words and special symbols to accurately transcribe sign language expressions, including gestures, non-manual features, etc. Its purpose is to reveal and record the unique grammatical structure of sign language. This annotation requires people familiar with both sign language and glossing conventions.} serve as a written representation of sign language, providing a high degree of semantic alignment. Currently, many studies focus on Gloss-based Sign Language Translation (GB-SLT). Camgoz \textit{et al}. \cite{camgoz2018neural} employs a sign language recognition model to generate gloss, which is then processed to produce the final translation. Alternatively, Necati \textit{et al}. \cite{camgoz2020sign} proposed using gloss annotations only during the training phase as a constraint to guide the model, helping it better understand sign language videos and improving its ability to process complex sign language information. However, although gloss can enhance sign language translation performance, its annotation process requires specialized professionals and is highly time-consuming. To reduce the annotation cost of gloss labels, Gloss-free Sign Language Translation (GF-SLT) methods \cite{zhou2023gloss, yin2023gloss}  eliminate the use of gloss as a constraint and directly generate text translations from sign language videos. However, due to the absence of gloss constraints, the accuracy of GF-SLT remains lower than that of GB-SLT.

\begin{figure} \centering
        \vspace{-20pt}
	\includegraphics[scale=0.65]{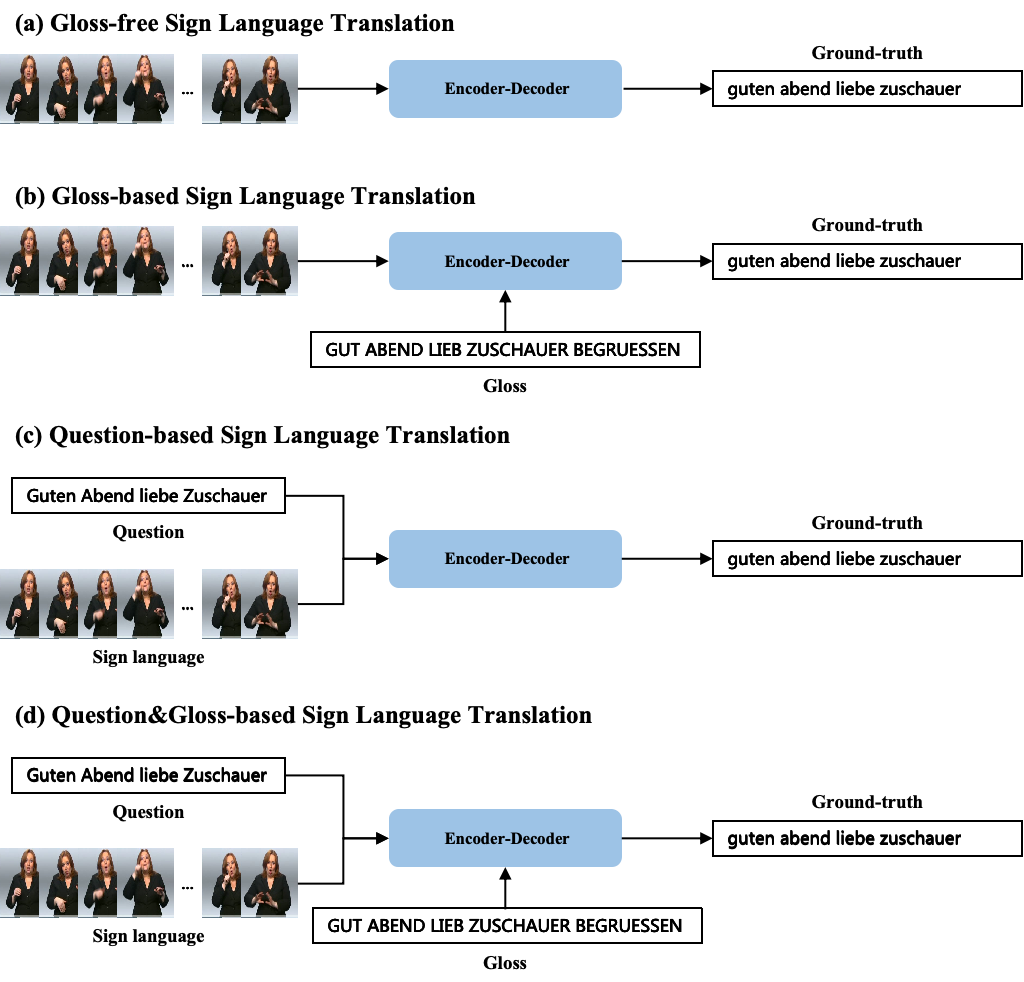}
	\caption{Distinction between Gloss-free Sign Language Translation, Gloss-based Sign Language Translation, Question-based Sign Language Translation and Question\&Gloss-based Sign Language Translation Tasks.} 
	\label{fig:definition}
	\vspace{-10pt}
\end{figure}

Actually, in honest communication, conversations between deaf people and hearing people usually contain contextual information, which can serve as additional clues to reduce ambiguity in translation. To leverage such contextual clues, Gao \textit{et al}. \cite{gao2024overcoming} proposed Question\&Gloss-based Sign Language Translation (QG-SLT), aiming to improve translation quality by incorporating relevant contextual information. Considering that QG-SLT still incurs a high annotation cost due to its reliance on glosses, this paper proposes a new task: Question-based Sign Language Translation (QB-SLT) for the first time. This task utilizes naturally occurring questions as auxiliary information, thereby eliminating the need for gloss annotations. The distinctions between QB-SLT and existing tasks such as GB-SLT, GF-SLT, and QG-SLT are illustrated in Figure \ref{fig:definition}. Building on the QB-SLT task, this paper proposes a cross-modality Self-supervised Learning with Sigmoid Self-attention Weighting (SSL-SSAW) fusion for sign language translation, which leverages questions as auxiliary information in place of traditional gloss to enhance translation accuracy.

\begin{figure}
	\vspace{-18pt}
	\subfloat[]{\includegraphics[width=0.5\textwidth]{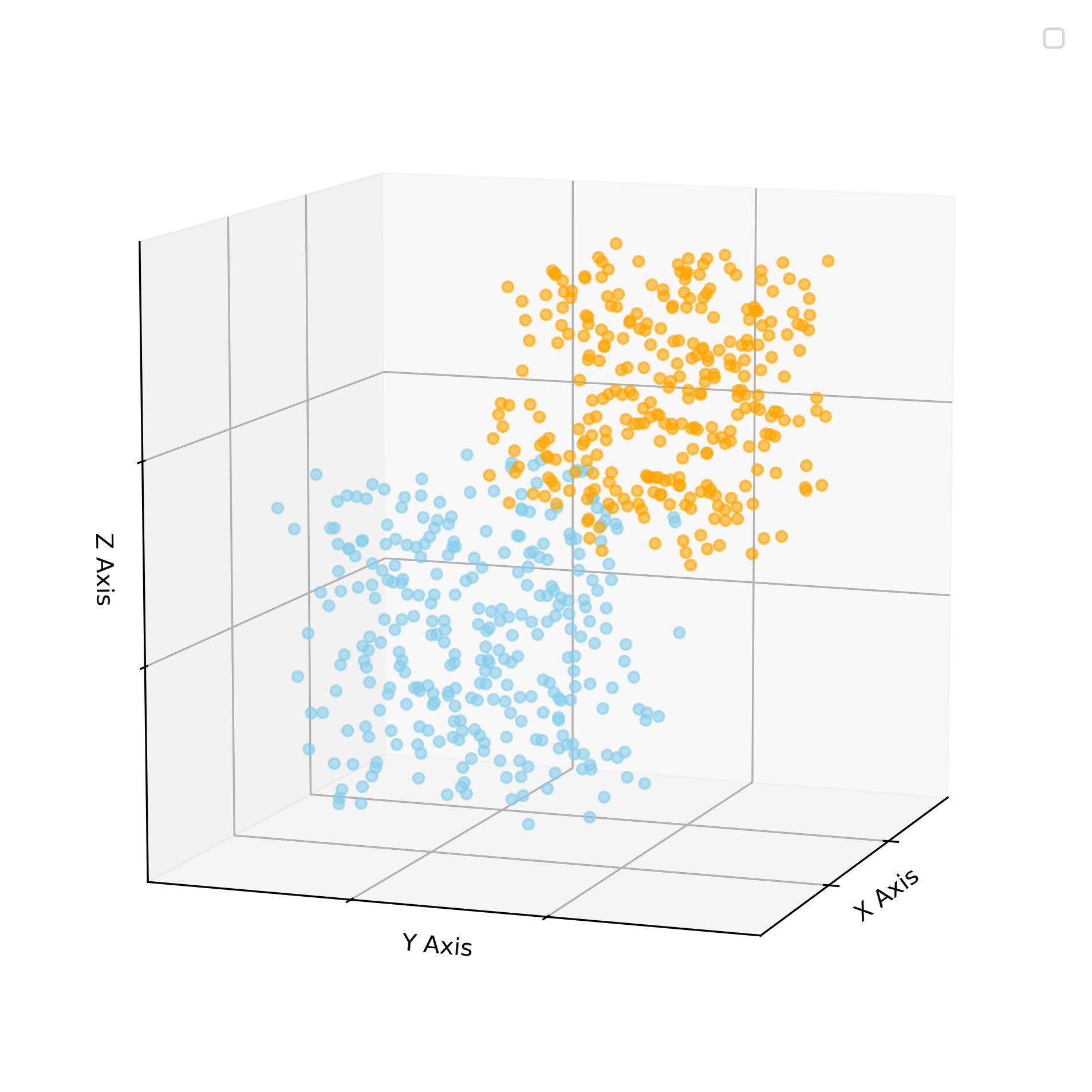}}
	\subfloat[]{\includegraphics[width=0.5\textwidth]{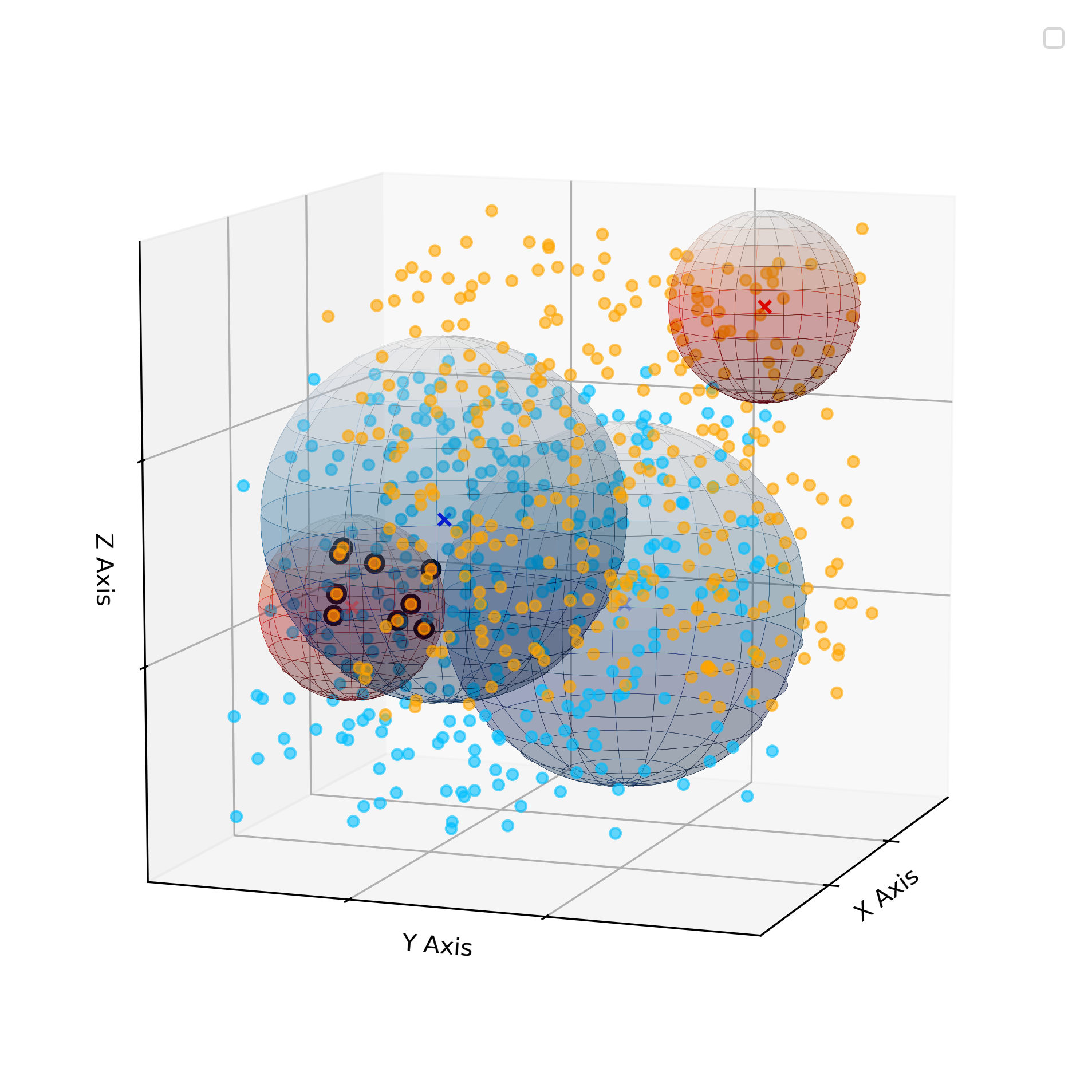}}
	\vspace{-8pt}
	
	\caption{A schematic diagram of the three-dimensional distribution of multimodality features. Orange dots represent text features, while blue dots represent video features. The orange sphere denotes the response space for dialog text, and the blue sphere denotes the response space for sign language videos. Within the response space, text features are highlighted as orange dots with black edges.}
	%\vspace{-12pt}
	\label{fig:feature_distribution}
	
	%\vspace{0.2in}
\end{figure}

This paper focuses on three core challenges that affect the effectiveness of QB-SLT tasks: (i) As shown in Figure \ref{fig:feature_distribution} (a), the orange and blue dots denote the sign language representation and the question representation, respectively. Since QB-SLT involves two modalities, video and text, the data of these two modalities have differences in semantic distribution. Therefore, it is necessary to construct a unified shared feature space to achieve efficient collaborative representation of cross-modality features. (ii) As shown in Figure \ref{fig:feature_distribution} (b), the orange and blue spheres represent the response space of sign language representation and question representation, respectively. Since the purpose of QB-SLT is to obtain accurate sign language translation, and only part of the question information is helpful for translation, the model needs to adaptively filter key question information (orange sphere at the bottom left) and suppress irrelevant or noisy information (orange sphere at the top right). (iii) How to leverage the available known information to enhance the expressive capacity of the shared feature space, aiming to enable the model to effectively associate with the representation of the translated text within a smaller response space (the orange and blue spheres in Figure \ref{fig:feature_distribution} (b) have smaller radii).

QB-SLT involves two modalities: question text and sign language video. Mikolov \textit{et al}.\cite{mikolov2013efficient} has demonstrated that in natural language processing tasks, models can learn to construct semantic relationships between words through large-scale data. Similarly, as a natural language, sign language also exhibits inherent semantic relationships between its vocabulary. Inspired by the success of CLIP \cite{radford2021learning}, we employ contrastive learning to construct a shared feature space for videos and text, allowing sign language representations to inherit semantic relationships from text representations. This ensures that question text and sign language video representations are computed within the same semantic space.

Questions often contain irrelevant words as auxiliary information, and the model should reduce its focus on such information. Suppose the same processing is applied to every word in the question during feature extraction. In that case, the model may learn excessive noisy information, which could impact its generalization ability and even disrupt the dependencies within the time series. To address this issue, we propose a Sigmoid Self-attention Weighting (SSAW) fusion module. The introduction of the sigmoid function into the fusion process allows the model to independently assign weights to each feature and dynamically adjust the importance of question text features and sign language video features for each frame. This process not only strengthens the correlation between modalities but also prevents the accumulation of noisy information.

Unlike other text-visual tasks \cite{chen2022pali, he2024vlab, wang2024qwen2, gao2024linvt, he2024ma}, sign language itself is a natural language. The process of sign language translation involves mapping sign language videos to a latent feature space and generating the corresponding text through a decoder. This process is similar to the self-supervised learning approach used in text autoencoding. Furthermore, during the pre-training phase, a shared feature space for video and text has already been constructed, which helps achieve efficient transfer learning and knowledge sharing between the sign language translation task and the text autoencoding task. Based on the unity of these two tasks, in order to fully leverage the available question information, this paper introduces a self-supervised text autoencoding task for the question text to enhance the model's ability to generalize and establish contextual relationships, making it more stable and efficient in the sign language translation task.

The major contributions are summerized:

%\begin{hangparas}{1.1em}{1}
\begin{itemize}
	\item This paper substitutes low-cost labeled question for gloss as auxiliary information in sign language translation, achieving or even surpassing the accuracy of models that use glosses for translation.
	
	\vspace{7pt}
	
	\item We propose Sigmoid Self-attention Weighting fusion to effectively integrate question text and sign language video features, selecting effective features for sign language translation.
	
	\vspace{7pt}
	
	\item By fully leveraging known question text for self-supervised learning, we enhance the model's ability to establish contextual relationships and improve its generalization capability.
\end{itemize}
%\end{hangparas}

\section{Related Work}
\subsection{Gloss-based Sign Language Translation}
For GB-SLT methods, an important characteristic is the direct or indirect use of gloss annotations from sign language to enhance the performance of sign language video encoders. These methods typically employ Connectionist Temporal Classification (CTC) \cite{graves2006connectionist} loss to perform sign language recognition. Necati \textit{et al}. \cite{camgoz2020sign} were the first to propose a Transformer-based encoder-decoder framework for the joint optimization of sign language recognition and translation. Zhou \textit{et al}. \cite{zhou2021spatial} introduced spatiotemporal modeling of multiple cues in sign language, using segment-level attention mechanisms to improve translation quality, achieving leading performance on several benchmark datasets.  Zhang \textit{et al}. \cite{zhang2023sltunet} proposed a unified framework for jointly training multiple SLT-related tasks to reduce the gap between the sign language and text modalities. Yao \textit{et al}. \cite{yao2023sign} proposed an iterative prototype mechanism that enables the model to progressively refine and enhance its semantic representation capabilities through repeated iterations, effectively emulating the human cognitive process of rereading and deepening understanding of sentences. Hu \textit{et al}. \cite{hu2023signbert+} reduced dependence on large-scale annotated data by introducing self-supervised pre-training of hand models. Chen \textit{et al}. \cite{chen2022simple} proposed a simple and effective transfer learning baseline by decomposing the sign language translation task into visual tasks (Sign2Gloss) and linguistic tasks (Gloss2Text), each of which is pre-trained using transfer learning to reduce the reliance on large-scale parallel data. Building on this, Chen \textit{et al}. \cite{chen2022two} introduced keypoint information and proposed a two-stream network that reduces visual redundancy and enhances the model's ability to extract significant information from sign language videos.

\subsection{Gloss-free Sign Language Translation}
Providing gloss annotations for sign language translation datasets is a resource-consuming and time-consuming process. Therefore, GF-SLT methods that do not rely on gloss annotations have become a promising alternative, offering greater generality. Necati \textit{et al}. \cite{camgoz2018neural} pioneered sign language translation as an NMT task, employing CNN-based encoders with attention mechanisms to map videos to spoken language. Subsequent work by Zhao \textit{et al}. \cite{zhao2021conditional} introduced a three-stage framework with word existence verification, conditional sentence generation, and cross-modality reranking to enhance lexical alignment and semantic consistency. Zhou \textit{et al}. \cite{zhou2023gloss} leveraged vision-language pretraining (VLP) combining CLIP and masked self-supervision. Attention mechanisms have also been refined for sign language translation. Thang \textit{et al}. \cite{luong2015effective} proposed local and global attention for context-aware decoding, Yin \textit{et al}. \cite{yin2023gloss} designed gloss attention to adjust frame-level attention ranges dynamically, and Li \textit{et al}. \cite{li2020tspnet} introduced hierarchical multi-scale attention via temporal semantic pyramids. With the advancement of large language models (LLMs \cite{vaswani2017attention}), many sign language GF-SLT methods that integrate LLMs have also achieved remarkable success. Gong \textit{et al}. \cite{gong2024llms} introduced the SignLLM framework, which discretizes sign language videos into hierarchical linguistic tokens (from character-level to word-level) and employs distribution alignment loss to bridge the semantic gap between modalities. To address annotation-free scenarios, Chen \textit{et al}. \cite{chen2024factorized} proposed a decomposed training strategy, FLa-LLM, which separates the pretraining of the visual encoder from the fine-tuning of the LLM to prevent representation degradation in joint optimization. 

\subsection{Question\&Gloss-base Sign Language Translation}
To better leverage real-world dialogues between deaf and hearing people, Gao \textit{et al}. \cite{gao2024overcoming} first proposed the Question\&Gloss-based Sign Language Translation (QG-SLT) task. They introduced the Gloss-Bridged Translator (GBT) model, which utilizes questions as auxiliary inputs and glosses as auxiliary supervision to significantly enhance model performance. Furthermore, they manually annotated questions to construct a rich Question-Driven Sign Language dataset for training and evaluating the model.

Current sign language translation (SLT) approaches mainly focus on three categories: GB-SLT, which uses gloss annotations to improve accuracy but requires costly professional labeling. GF-SLT, which operates without gloss but suffers from lower performance due to an unconstrained translation space. QG-SLT combines gloss and question annotations at the same time, and performs well in accuracy, but still cannot get rid of the dependence on high-cost annotations. This paper introduces Question-based SLT (QB-SLT), a novel task that only leverages naturally occurring question context as auxiliary information. Unlike gloss annotations, questions can be easily generated from spoken text labels, making QB-SLT both practical and effective for real-world communication scenarios between deaf people and hearing people.

\section{Our Approach}
\label{sec:approach} 

\subsection{Overview}

\begin{figure*} \centering
	\includegraphics[scale=0.48]{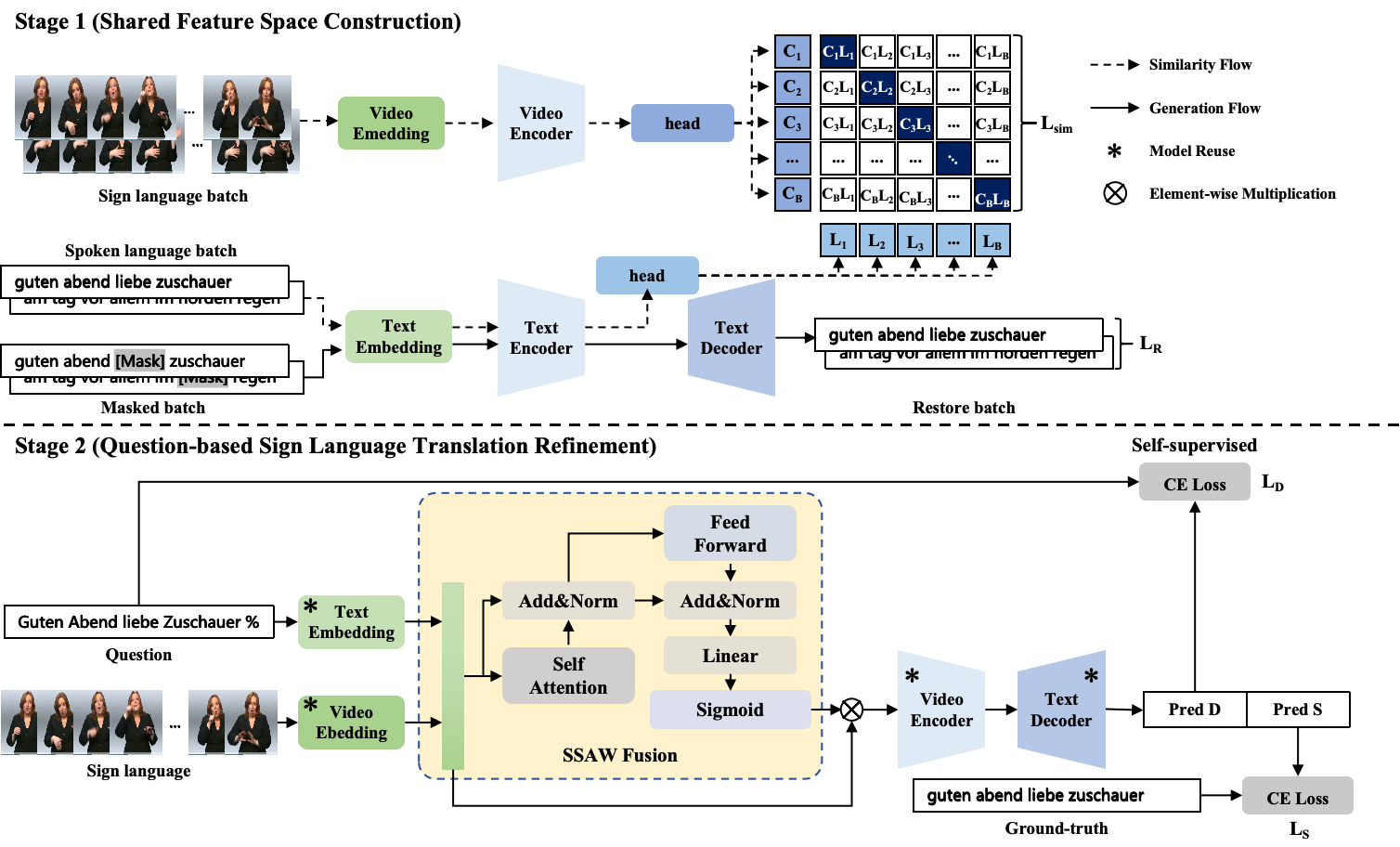}
	\caption{Overview of SSL-SSAW. The \textbf{dashed arrows} indicate the data flow for computing similarities. The \textbf{solid arrows} indicate the data flow for generating translations. \textbf{*} indicates model parameter reuse. \textbf{\textcircled{$\times$}} denotes the element-wise product.} 
	\label{fig:model}

    \vspace{-10pt}
\end{figure*}

Our proposed SSL-SSAW framework (Fig. \ref{fig:model}) enhances sign language translation by integrating question context and employing a two-stage training process. In Stage 1 (Shared Feature Space Construction), we employ contrastive learning to align video representations \{$C_1, \cdots, C_B$\} with corresponding text embeddings \{$L_1, \cdots, L_B$\}, while using Masked Language Modeling \cite{devlin2019bert} to preserve textual semantics. In Stage 2 (Question-based Sign Language Translation Refinement), we reuse pretrained encoders/decoders and introduce our novel SSAW module, which dynamically selects relevant question tokens to assist translation. Moreover, we integrate self-supervised learning to improve the encoder and decoder's ability to capture latent representations. Specifically, the question text is used both as input and as a label to guide the model in learning richer semantic features.

\subsection{Shared Feature Space Construction}
\label{subsec:pre}
QB-SLT involves data from both text and video modalities. Achieving cross-modality interaction requires that features from both modalities exist within a unified shared feature space. Inspired by CLIP \cite{radford2021learning} and GFSLT-VLP \cite{zhou2023gloss}, we employ contrastive learning to increase the similarity between text and video representations, which not only unifies the shared feature space but also enables video representations to inherit the semantic relationships from text representations. As a result, by integrating question text, QB-SLT better establishes the connection between video and question features.

Specifically, the input consists of sign language video sequence $ V=\{ x_{v}^{1}, x_{v}^{2}, \\ \cdots, x_{v}^{N}\}$ and the corresponding translation text sequences $ S=\left\{ x_{s}^{1},x_{s}^{2},\cdots ,x_{s}^{M}\right\}$. We use Video Embedding and Video Encoder to extract visual features $f_{VE} = \psi_{VE}(V)$. We utilize $\psi_{TE}$ (including Text Embedding and Text Encoder) to extract textual features $f_{SE} = \psi_{TE}(S)$. Following Devlin \textit{et al}.. \cite{devlin2019bert}, the feature corresponding to the [EOS] or [CLS] token can effectively represent the semantic content of the entire sequence. Therefore, we use the visual feature at the [CLS] token $f_{VE,N}$ and the text feature at the [EOS] token $f_{SE,M}$ to compute the subsequent similarity.
The extracted features $f_{VE,N}$ and $f_{SE,M}$ are then passed through a Linear layer, followed by the computation of similarity loss between the video and text representations using a symmetric cross-entropy loss function, as follows:

\begin{align}
	&I_s = {\rm Linear}(f_{SE,M}), I_v = {\rm Linear}(f_{VE,N}), \\
	&L_{sim} = -\frac{1}{2} \sum(V \log(I_{s})) + \sum(S \log(I_{v})).  \notag
\end{align}

Additionally, this paper masks individual words in the spoken text and regenerates them using the Text Embedding-Encoder-Decoder. This ensures that, while constructing a shared feature space, the features extracted by the model are aligned with the requirements of the sign language translation task. This process is illustrated in Equation \ref{eq:mask}.

\begin{equation}
	\min_{\Theta} \frac{1}{q} \sum_{i=1}^{q} L_R ( \psi_{TD} ( \psi_{TE}^* ( \tilde{x}_{s}^{i} ) ), x_{s}^{i}),
    \label{eq:mask}
\end{equation}
where $q$ is the number of training samples, $L_R$ is the Cross Entropy loss function, $\psi_{TD}$ is the Text Decoder, $\psi_{TE}^*$ represents the shared parameters with the Text Encoder for calculating similarity. $\tilde{x}_{s}^{i}$ represents the masked word.

\subsection{Question-based Sign Language Translation Refinement}

To effectively leverage question information for sign language translation while mitigating the impact of noisy content in the question, this paper proposes a Sigmoid Self-Attention Weighting (SSAW) Fusion module and a Self-supervised Learning (SSL) strategy that integrates question context for sign language translation.

\textbf{Sigmoid Self-attention Weighting Fusion.} In this paper, we introduce the question sequence as auxiliary information for the sign language translation. However, this may also introduce redundant or even noisy information that could negatively impact the translation process. Additionally, sign language videos contain transition frames and similar redundant frames, making it challenging to distinguish the contribution of each frame to the translation task using only temporal feature extraction methods.

To address this, we propose the Sigmoid Self-attention Weighting (SSAW) fusion module, which integrates question and sign language video features. The self-attention mechanism captures long-range dependencies, dynamically focusing on key elements in both sequences to strengthen contextual relationships. 
Then, the nonlinear mapping is performed through the Feed Forward module to further improve the representation capability of the sequence features. To improve feature sparsity, we amplify the importance of critical features while preserving the potential influence of minor ones. The sigmoid activation function is employed to independently assess the importance of each feature. By facilitating better cross-modality fusion of text and sign language, this approach enhances both the accuracy and fluency of sign language translation.

Specifically, the features $f_{e,\rm{DE}}$ extracted from the question via Text Embedding and the features $f_{e,\rm{VE}}$ extracted from the sign language via Video Embedding are concatenated along the temporal dimension to form the combined feature $f_{c}$.  For the combined feature, we calculate the query $Q_{c}$, key $K_{c}$, and value $V_{c}$ using learnable weights $W_{Q}$, $W_{K}$ and $W_{V}$. The self-attention (SA) mechanism is applied by computing the aggregated feature representation $f_{a}$. After passing through the self-attention, residual connections and Layer Normalization (LN) are applied to the features. 
\begin{align}
	&f_{c} = {\rm{Concat}}(f_{e,\rm{DE}}, f_{e,\rm{VE}}), \notag \\
	&Q_{c} = f_{c}W_{Q}, K_{c} = f_{c}W_{K}, V_{c} = f_{c}W_{V},  \\
	&f_{a} = {\rm{LN}}((f_{c} + {\rm{SA}}(Q_{c}, K_{c}, V_{c})). \notag 
\end{align}

The feature $f_{a}$ is then processed through two linear layers for feature extraction. After passing through the first linear layer, a ReLU activation function is applied for nonlinear transformation. Similarly, after each pass through the Feed Forward Network (FFN), residual connections and LN are applied to the features to ensure their stability.
\vspace{-5pt}

\begin{equation}
	f_{f} ={\rm{LN}}( f_{a} + ({\rm{ReLU}}( f_{a}W_{L1} + b_{L1})W_{L2} + b_{L2})),
\end{equation}
where $W_{L1}$, $W_{L2}$, $b_{L1}$ and $b_{L2}$ represent the learnable parameters and biases of the first and second linear layers, respectively. Finally, the sigmoid ($\sigma$) function is applied to compute the weight score of $f_{f}$, which is then multiplied by the original combined feature $f_{c}$ to obtain the weighted joint feature $f_{j} = f_{c} \cdot \sigma(f_{f})$

\textbf{Self-supervised Sign Language Translation.} We employ a self-supervised learning strategy to enhance the model's ability to learn latent representations and improve the quality of translation generation. By using a question as both input and label, the model autonomously learns richer feature representations without relying on manually annotated data. This self-supervised mechanism not only significantly enhances the model's generalization ability and robustness but also enables it to efficiently capture complex patterns across both spoken language and sign language modalities.

Specifically, for the question sequence $D$, each current prediction is based on the encoder hidden states $h_{j}$ and the question encoding before the current time step $z_{i}^{d} = {\rm softmax}(W \psi_{\rm TD}^{*}(h_j, D_{1 : i-1})+b)$, where $W$ and $b$ are the learnable parameters and bias of a Linear layer. $D_{i-1} = (d_1, \cdots, d_{i-1})$, where the first character $d_1$ is start of sentence token $<$BOS$>$. $ \psi_{\rm TD}^{*}$ is the text decoder mentioned in Section \ref{subsec:pre}, which aligns the video modality and text modality through pre-training. 

The purpose of utilizing question sequence training is to enhance the model's capabilities in text understanding, video understanding, text generation, and generalization. Since the question sequence is known, we employ a self-supervised learning approach for training, minimizing the discrepancy between the model's prediction and the input question sequence. The loss function is given by, where $p(\cdot|\cdot)$ denotes the conditional probability:

\begin{equation}
	L_{D} = -\sum_{i=1}^{M}{\rm log}(p(d_{i}|z_{i}^{d})).
\end{equation}

For the spoken text sequence $S$, each current prediction is based on the encoder hidden states $h_{j}$, the complete question sequence encoding $D_{M}$, and the encoding of the spoken text sequence before the current moment $z_{k}^{s} = {\rm softmax}(W \psi_{\rm TD}^{*}(h_j, D_{M}, S_{k-1})+b)$, where  $S_{k-1} = (s_1, \cdots, s_{k-1})$. The goal of training on the spoken text sequence is to minimize the difference between the model's prediction and the actual spoken text label. The loss function can be expressed as:

\begin{equation}
	L_{S} = - \sum_{k=1}^{N}{\rm log}(p(s_{k}|z_{k}^{s})).
\end{equation}

Since the modality alignment has been completed in Section \ref{subsec:pre}, it has enhanced the consistency between the tasks of question self-supervised learning and sign language translation, while also achieving unity in the multimodality feature space. As a result, we can effectively combine the loss functions of both tasks:

\begin{equation}
	L_{total} = L_{D} + L_{S}.
\end{equation}

\section{Experimental Analysis}
% The network architecture presented in \textbf{Section 3} has been tested and validated on the BR-CSL dataset. In the subsequent sections, we provide detailed information about the dataset, model hyperparameters, and hardware setup employed in our experiments. Meanwhile, the experimental results are quantitatively analyzed.

\subsection{Dataset}To evaluate the effectiveness of QB-SLT, we conducted experiments using the PHOENIX-2014T-QA \cite{gao2024overcoming} and CSL-Daily-QA \cite{gao2024overcoming} datasets, both of which were constructed based on the publicly available PHOENIX-2014T \cite{camgoz2018neural} and CSL-Daily \cite{zhou2021improving} datasets. The PHOENIX-14T dataset, recorded by 9 signers, contains 7,096 training videos, 519 validation videos, and 642 test videos. The CSL-Daily dataset was recorded by 10 signers and consists of 18,401 training videos, 1,077 validation videos, and 1,176 test videos. Both datasets were constructed through offline interpretation rather than real-time translation.
PHOENIX-2014T focuses on the German daily news and weather forecast program, with the sign language primarily covering weather-related topics. And CSL-Daily focuses on topics related to people's daily lives (e.g., travel, shopping, medical care). In these datasets, both spoken language and sign language are prepared in advance using scripts.

Based on these datasets, \cite{gao2024overcoming} manually annotated relevant questions for each video, resulting in the creation of the PHOENIX-2014T-QA and CSL-Daily-QA datasets. These questions are formulated as natural language sentences and are contextually linked to the corresponding sign language videos. To ensure the quality and validity of the annotations, all questions were verified through cross-validation.

\subsection{Experimental Setup}
For the proposed SSL-SSAW, we use Conv2D to extract spatial features from video frames and use Conv1D-BN-Relu-Maxpooling to capture short-term features of the video. These two parts, as video embedding, are used to extract the video representation. And Text Embedding directly uses the Embedding of pre-trained mBART \cite{liu2020multilingual}. Both Text Encoder, Text Decoder and Video Encoder are initialized with the pre-trained parameters of the mBART model. The proposed method is implemented in PyTorch and the experiments are carried out using a NVIDIA GeForce RTX 3090-GPU.

To evaluate the performance of the sign language translation model, we choose BLEU-n \cite{papineni2002bleu} and ROUGE \cite{lin2004rouge} as evaluation metrics. BLEU-n is a metric for evaluating the quality of machine translation by measuring the overlap of n-grams between the machine-generated translation and the reference translation. ROUGE evaluates content coverage by measuring the recall of n-gram co-occurrence and the longest common subsequence between the generated text and the reference text.

\subsection{Experimental Results}
\textbf{Comparison experiments on the PHOENIX-2014T-QA dataset.} To evaluate model performance, we systematically compare state-of-the-art (SOTA) models for GF-SLT, GB-SLT, QG-SLT, and QB-SLT tasks on the PHOENIX-2014T-QA dataset. As shown in Table \ref{tab:p}, the proposed SSL-SSAW, which leverages easily accessible and low-cost question annotations, achieves significant improvements over the current GB-SLT model (TS-SLT), with gains of 6.47 in BLEU-4 and 6.68 in ROUGE. Compared to the best-performing model in QB-SLT (QB-GFSLT-VLP), SSL-SSAW yields improvements of 9.67 in BLEU-4 and 10.84 in ROUGE. Additionally, our model outperforms the representative QG-SLT model (GBT) by 1.11 in BLEU-4 and 0.66 in ROUGE, further validating the effectiveness of our approach.

A comparison with QB-SLT models reveals that, even within a shared feature space, semantic discrepancies persist between different modalities. Moreover, although dialogue information can serve as auxiliary input, it often contains noise that may hinder the sign language translation process and reduce accuracy. However, in the PHOENIX-2014T-QA dataset, the questions are highly relevant to the video content (e.g., weather forecasts), which helps the model to capture semantic relationships between questions and translations better. Based on this, our SSL-SSAW effectively integrates question information and enhances feature representation and generation capabilities through self-supervised learning, ultimately leading to substantial improvements in sign language translation performance.

\begin{table}[H]
	\centering
	\caption{Comparison with other SOTA methods on the PHOENIX-2014T-QA dataset.}\label{tab:p}
	\resizebox{0.9\columnwidth}{!}{

		\begin{tabular}{  c | c c c c c }
			\Xhline{2pt}
			\textbf{Method} 				    & \textbf{B1}$\uparrow$ & \textbf{B2}$\uparrow$ & \textbf{B3}$\uparrow$ & \textbf{B4}$\uparrow$ & \textbf{ROUGE}$\uparrow$  \\
			\hline
			\multicolumn{6}{c}{ \cellcolor{gray!30} Gloss-free SLT (GF-SLT)} \\
			\hline
			NSLT \cite{camgoz2018neural} & 27.10 & 15.61 & 10.82 & 8.35 & 29.70  \\
			
			NSLT+Bahdanau \cite{camgoz2018neural, bahdanau2014neural} & 32.24 & 19.03 & 12.83 & 9.58 & 31.80  \\
			
			NSLT+Luong \cite{camgoz2018neural, luong2015effective} & 29.86 & 17.52 & 11.96 & 9.00 & 30.70 \\
			
			TSPNet \cite{li2020tspnet} & 36.10 & 23.12 & 16.88 & 13.41 & 34.96 \\
			
			CSGCR \cite{zhao2021conditional} & 36.71 & 25.40 & 18.86 & 15.18 & 38.85 \\
			
			GASLT \cite{yin2023gloss} & 39.07 & 26.74 & 21.86 & 15.74 & 39.86 \\
			
			GFSLT-VLP \cite{zhou2023gloss} & 43.71& 33.18 & 26.11 & 21.44 & 42.49 \\
			
			\hline
			\multicolumn{6}{c}{ \cellcolor{gray!30} Gloss-based SLT (GB-SLT)} \\
			\hline
			
			STMC-T \cite{zhou2021spatial} & 46.98 & 36.09 & 28.70 & 23.65 & 46.65 \\
			
			BTT+SignBT \cite{zhou2021improving} & 50.80 & 37.75 & 29.72 & 24.32 & 49.54 \\
			
			MMTLB \cite{chen2022simple} & 53.97 & 41.75 & 33.84 & 28.39 & 52.65 \\
			
			TS-SLT \cite{chen2022two} & 54.90 & 42.43 & 34.46 & 28.95 & 53.48 \\

			\hline
			\multicolumn{6}{c}{ \cellcolor{gray!30} Question\&Gloss-based SLT (QG-SLT)} \\
			\hline
			
			GBT \cite{gao2024overcoming} & 60.85 & 49.00 & 40.50 & 34.31 & 59.50 \\
			
			% SCOPE & \textbf{61.74} & 49.22 & 39.61 & 32.84 & 60.06 \\
			
			\hline
			\multicolumn{6}{c}{ \cellcolor{gray!30} Question-based SLT (QB-SLT)} \\
			\hline
			
			QB-SLRTt \cite{camgoz2020sign} & 38.04 & 25.53 & 18.70 & 14.49 & 36.24 \\
			
			QB-GASLT \cite{yin2023gloss} & 40.21 & 28.77 & 21.91 & 17.37 & 42.64 \\
			
			QB-GFSLT-VLP \cite{zhou2023gloss} & 48.98 & 38.69 & 31.18 & 25.75 & 49.32 \\
			
			\hline 
			\textbf{SSL-SSAW(Ours)} & \textbf{61.09} & \textbf{49.95} & \textbf{41.55} & \textbf{35.42 } & \textbf{60.16} \\

			\Xhline{2pt}

		\end{tabular}
		}
	\vspace{-10pt}
\end{table}

\textbf{Comparison experiments on the CSL-Daily-QA dataset.} To evaluate the generalization and robustness of our model, we conducted comparative experiments on the CSL-Daily-QA dataset. As shown in Table 2, the proposed SSL-SSAW model outperforms the current SOTA model QB-GFSLT-VLP (QB-SLT) across all metrics. Compared to GB-SLT and QG-SLT models, which rely on gloss annotations that are difficult to annotate, SSL-SSAW shows slightly lower performance than TS-SLT and GBT on BLEU-1 and ROUGE metrics that emphasize word-level matches. This is primarily because TS-SLT and GBT use gloss annotations that closely follow the word order of sign language videos, providing strong supervision and enabling more precise word-level alignment. However, in terms of BLEU-4, which places more emphasis on word sequence coherence and semantic relationships, SSL-SSAW achieves significant improvements of 8.47 and 0.86 over TS-SLT and GBT, respectively, demonstrating superior semantic comprehension.

\begin{table}
	\centering
	\vspace{-15pt}
        \caption{Comparison with other SOTA methods on the CSL-Daily-QA dataset.}\label{tab:CSL-Daily}
	\resizebox{0.84\columnwidth}{!}{
		\begin{tabular}{  c | c c c c c }
			\Xhline{2pt}
			\textbf{Method} 				    & \textbf{B1}$\uparrow$ & \textbf{B2}$\uparrow$ & \textbf{B3}$\uparrow$ & \textbf{B4}$\uparrow$ & \textbf{ROUGE}$\uparrow$  \\
			\hline
			\multicolumn{6}{c}{ \cellcolor{gray!30} Gloss-free SLT  (GF-SLT)} \\
			\hline
			SLRTt \cite{camgoz2020sign} &20.00 & 9.11 & 4.93 & 3.03 & 19.67  \\
			
			GASLT \cite{yin2023gloss} & 19.90 & 9.94 & 5.98 & 4.07 & 20.35  \\
			
			NSLT+Luong \cite{camgoz2018neural, luong2015effective} & 34.16 & 19.57 & 11.84 & 7.56 & 34.54 \\
			
			GFSLT-VLP \cite{zhou2023gloss} & 39.37 & 24.93 & 16.26 & 11.00 & 36.44  \\
			
			\hline
			\multicolumn{6}{c}{ \cellcolor{gray!30} Gloss-based SLT (GB-SLT)} \\
			\hline
			
			BTT+SignBT \cite{zhou2021improving} & 51.42 & 37.26 & 27.76 & 21.34 & 49.31 \\
			
			MMTLB \cite{chen2022simple} & 53.31 & 40.41 & 30.87 & 23.92 & 53.25 \\
			
			TS-SLT \cite{chen2022two} & 55.44 & 42.59 & 32.87 & 25.79 & 55.72 \\
			
			\hline
			\multicolumn{6}{c}{ \cellcolor{gray!30} Question\&Gloss-based SLT (QG-SLT)} \\
			\hline
			
			GBT \cite{gao2024overcoming} & \textbf{57.37} & \textbf{46.51} & 39.06 & 33.40 & \textbf{57.81} \\
			
			% SCOPE & \textbf{60.48} & \textbf{49.61} & 40.01 & 32.08 & \textbf{60.68} \\
			
			\hline
			\multicolumn{6}{c}{ \cellcolor{gray!30} Question-based SLT (QB-SLT)} \\
			\hline
			QB-SLRTt \cite{camgoz2020sign} & 17.37 & 8.17 & 3.35 & 2.71 & 16.38 \\
			
			QB-GASLT \cite{yin2023gloss} & 21.97 &12.37 & 9.38 & 6.87 & 24.75 \\
			
			QB-GFSLT-VLP \cite{zhou2023gloss} & 28.10 & 18.67 & 12.27 & 8.37 & 27.21 \\

			\hline 
			\textbf{SSL-SSAW(Ours)} & 51.67 & 45.88 & \textbf{39.14} & \textbf{34.26} & 52.59 \\
			
			\Xhline{2pt}

		\end{tabular}
	}
	\vspace{-10pt}
\end{table}

\textbf{Ablation study.} Stage 1 of this study focuses on establishing semantic alignment between sign language videos and their corresponding translated text. Since the lengths of the question text sequences and sign language video sequences differ, we first perform sequence length alignment, followed by semantic matching. In our method, both the sign language video and the translated text are encoded into a 1D vector representation. This design is motivated by two main reasons: (1) the differing lengths make it difficult to achieve strong, frame-level or token-level alignment between sequences; and (2) contrastive learning aims to capture the overall semantic correspondence between the entire video and its translation, emphasizing global consistency over local alignment.

To evaluate the effectiveness of different alignment strategies, we conducted ablation experiments using two approaches: one maps the longer video sequence to match the length of the text sequence, while the other applies bilinear interpolation to expand the shorter text sequence to align with the video sequence. The comparison results of these two approaches are shown in Table \ref{tab:align}.

\begin{table}[H]
	\centering
	\caption{Comparison of different alignment strategies' impact on QB-SLT performance on the PHOENIX-2014T-QA dataset.}\label{tab:align}
	\resizebox{0.95\columnwidth}{!}{

		\begin{tabular}{  c | c c c c c }
			\Xhline{2pt}
			\textbf{Method} 				    & \textbf{B1}$\uparrow$ & \textbf{B2}$\uparrow$ & \textbf{B3}$\uparrow$ & \textbf{B4}$\uparrow$ & \textbf{ROUGE}$\uparrow$  \\
			\hline
			$|$Text$|$ $\to$ $|$Video$|$ & 56.24 & 45.86 & 38.33 & 32.04 & 56.03 \\
			
			$|$Video$|$ $\to$ $|$Text$|$ & 58.36 & 47.65 & 39.71 & 33.92 & 57.96 \\
			
			\textbf{$|$Video$|$, $|$Text$|$ $\to$  1 (Ours)} & \textbf{61.09} & \textbf{49.95} & \textbf{41.55} & \textbf{35.42} & \textbf{60.16} \\

			\Xhline{2pt}

		\end{tabular}
		}
	\vspace{-10pt}
\end{table}
$|\cdot|$ denotes the sequence length, and $\to$ indicates sequence alignment. Experimental results show that encoding both the sign language video and the translated text into a single unified vector is more effective for modeling their overall semantic consistency, as sequence alignment may introduce semantic ambiguity.

As mentioned in Section \ref{sec:approach}, we introduce question information into the sign language translation task and propose the SSAW module and an SSL strategy to address the issues of cross-modality fusion and the effective use of question information. We conducted ablation experiments on the PHOENIX-2014T-QA dataset to verify the impact of each module and strategy. The experimental results are shown in Table \ref{tab:a}, where QB refers to the approach of simply concatenating question features with video features, followed by training and testing using the baseline model. The results demonstrate that, on the PHOENIX-2014T-QA dataset, introducing question information effectively improves the model's sign language translation performance, with improvements of 4.31 and 6.83 in BLEU-4 and ROUGE, respectively.

\begin{table}[H]
	\centering
	\caption{Ablation results of the SSL-SSAW design on the PHOENIX-2014T and PHOENIX-2014T-QA dataset.}\label{tab:a}
	\resizebox{0.84\columnwidth}{!}{
		\begin{tabular}{ c | c | c | c c c c c }
			\Xhline{2pt}
			\textbf{QB} & \textbf{SSL} & \textbf{SSAW} & \textbf{B1}$\uparrow$ & \textbf{B2}$\uparrow$ & \textbf{B3}$\uparrow$ & \textbf{B4}$\uparrow$ & \textbf{ROUGE}$\uparrow$ \\
			\hline
			\ding{55} & \ding{55} & \ding{55} &  43.71 & 33.18 & 26.11 & 21.44 & 42.49		  \\
			
			\ding{51} & \ding{55} & \ding{55} & 48.98 & 38.69 & 31.18 & 25.75 & 49.32 \\
			
			\ding{51} & \ding{51} & \ding{55} & 51.05 & 40.50 & 33.92 & 27.69 & 51.23\\
			
			\ding{51} & \ding{55} &\ding{51}& 59.58 & 48.27 & 39.98 & 34.27 & 58.23\\
			
			\ding{51} & \ding{51} & \ding{51} & \textbf{61.09} & \textbf{49.95} & \textbf{41.55} & \textbf{35.42} & \textbf{60.16} \\
			\Xhline{2pt}

		\end{tabular}
	}
	\vspace{-10pt}
\end{table}

SSL and SSAW represent the methods of using questions for self-supervised training and the SSAW module, respectively. Since both frameworks are built upon the introduction of a question, all ablation experiments were based on the QB module. Experimental results show that simply introducing question information as auxiliary input leads to improvements of 4.31 in BLEU-4 and 6.83 in ROUGE scores. To further demonstrate the effectiveness of the SSL strategy and the SSAW module, the analysis compares the baseline QB model with versions augmented by SSL and SSAW. Applying the SSL strategy to the question text improves BLEU-4 by 1.94 and ROUGE by 1.91. Furthermore, leveraging the SSAW module for cross-modality fusion improves BLEU-4 by 8.52 and ROUGE by 8.91. Overall, this paper integrates questions as auxiliary information and integrates it with the SSL-SSAW framework, achieving significant improvements of 13.93 in BLEU-4 and 17.35 in ROUGE, demonstrating its effectiveness in enhancing sign language translation.

As shown in Table \ref{tab:c}, this paper also compares the impact of the proposed SSAW fusion method and other temporal feature fusion methods on cross-modality feature fusion. All methods in the table introduce question information and employ the SSL strategy. Specifically, the SSL-Concat method directly concatenates the features after Text Embedding and Video Embedding along the temporal dimension, which are then fed into the subsequent model for processing. The SSL-LSTM and SSL-TCN methods first concatenate the features from text and video embeddings along the temporal dimension, then model temporal dependencies using LSTM \cite{graves2012long} or TCN \cite{zhang2015character}, and finally feed the resulting features into the subsequent model.

\begin{table}[H]
	\centering
        \vspace{-10pt}
	\caption{Comparison of different fusion methods' impact on QB-SLT performance on the PHOENIX-2014T-QA dataset.}\label{tab:c}
	\resizebox{0.84\columnwidth}{!}{

		\begin{tabular}{  c | c c c c c }
			\Xhline{2pt}
			\textbf{Method} 				    & \textbf{B1}$\uparrow$ & \textbf{B2}$\uparrow$ & \textbf{B3}$\uparrow$ & \textbf{B4}$\uparrow$ & \textbf{ROUGE}$\uparrow$  \\
			\hline
			SSL-Concat & 48.98 & 38.69 & 31.18 & 25.75 & 49.32  \\
			
			SSL-LSTM & 54.41 & 45.73 & 39.04 & 34.08 & 56.31  \\
			
			SSL-TCN & 54.62 & 45.99 & 39.30 & 34.31 & 55.65  \\
			
			\textbf{SSL-SSAW(Ours)} & \textbf{61.09} & \textbf{49.95} & \textbf{41.55} & \textbf{35.42} & \textbf{60.16} \\

			\Xhline{2pt}

		\end{tabular}
	}
	\vspace{-10pt}
\end{table}

The experimental results indicate that the cross-modality fusion method has a significant impact on QB-SLT performance. As common temporal feature fusion methods, LSTM and TCN show improved accuracy compared to the simple Concat method. SSL-LSTM improves the BLEU-4 and ROUGE scores by 8.33 and 6.99, respectively, while SSL-TCN improves the BLEU-4 and ROUGE scores by 8.56 and 6.33, respectively. The proposed SSL-SSAW method not only focuses on feature fusion but also emphasizes the suppression of noisy information in the question information. Compared to SSL-Concat, SSL-SSAW improves BLEU-4 and ROUGE scores by 9.67 and 10.84, respectively.

As shown in Table \ref{tab:embedding}, this paper compares the robustness of different model architectures concerning various visual and text embedding modules. We conduct ablation experiments using Shufflenet \cite{zhang2018shufflenet} and ResNet \cite{he2016deep} as visual encoders, and mBERT \cite{devlin2019bert} and mBART as text encoders.

\begin{table}[H]
	\centering
	\caption{Comparison of different embedding modules' impact on QB-SLT performance on the PHOENIX-2014T-QA dataset.}\label{tab:embedding}
	\resizebox{0.92\columnwidth}{!}{

		\begin{tabular}{  c | c | c c c c c }
			\Xhline{2pt}
			\textbf{Video Emb} & 	\textbf{Text Emb} & \textbf{B1}$\uparrow$ & \textbf{B2}$\uparrow$ & \textbf{B3}$\uparrow$ & \textbf{B4}$\uparrow$ & \textbf{ROUGE}$\uparrow$  \\
			\hline
			ShuffleNet & mBERT & 54.18 & 44.36 & 34.89 & 29.80 & 50.08  \\
			
			ShuffleNet & mBART & 56.79 & 45.24 & 36.96& 30.92 & 51.85 \\
			
			ResNet & mBERT & 57.62 & 46.38 & 37.45 & 31.24 & 52.65  \\
			
			\textbf{ResNet} &\textbf{mBART} & \textbf{61.09} & \textbf{49.95} & \textbf{41.55} & \textbf{35.42} & \textbf{60.16} \\

			\Xhline{2pt}

		\end{tabular}
		}
	\vspace{-10pt}
\end{table}

Based on the results, mBART demonstrates superior performance as a text embedding module, while ResNet outperforms ShuffleNet in visual feature extraction. Consequently, this paper adopts ResNet as the visual embedding module and mBART as the text embedding module.

Incorporating a dialogue into sign language translation (SLT) tasks facilitates a more efficient video-to-text translation process. In interactions between hearing people and deaf people, an implicit contextual relationship often exists between spoken language and sign language. Such a question context provides valuable semantic cues that can enhance the model's understanding of sign language content and improve translation accuracy. To investigate the extent to which the model can leverage information from textual prompts, we conduct an ablation study involving three input configurations: using sign language video only, using question text only, and using both modalities simultaneously. The experimental results are presented in Table \ref{tab:modalities}.

\begin{table}[H]
	\centering
	\caption{Ablation results of the SSL-SSAW design on the PHOENIX-2014T-QA dataset.}\label{tab:modalities}
	\resizebox{0.81\columnwidth}{!}{
		\begin{tabular}{ c | c | c c c c c }
			\Xhline{2pt}
			\textbf{Question} & \textbf{Sign} & \textbf{B1}$\uparrow$ & \textbf{B2}$\uparrow$ & \textbf{B3}$\uparrow$ & \textbf{B4}$\uparrow$ & \textbf{ROUGE}$\uparrow$ \\
			\hline
			
			\ding{51} & \ding{55} & 46.82& 34.57 & 26.48 & 20.95 & 42.11 \\
			
			\ding{55} & \ding{51} & 45.30 & 34.76 & 28.03 & 23.41 & 44.75 \\
			
			\ding{51} & \ding{51} & \textbf{61.09} & \textbf{49.95} & \textbf{41.55} & \textbf{35.42} & \textbf{60.16} \\
			\Xhline{2pt}

		\end{tabular}
		}
	\vspace{-10pt}
\end{table}

Experimental results show that using only the question text as input yields lower performance compared to using both the question text and the sign language video together. In some metrics, it even performs worse than using only the sign language video. This demonstrates that the proposed SSL-SSAW can effectively extract useful information from both the question and the sign language input to improve translation accuracy, rather than relying solely on the easily learnable explicit mappings between question-answer pairs for prediction.

\begin{figure} [H]\centering
	\includegraphics[scale=0.56]{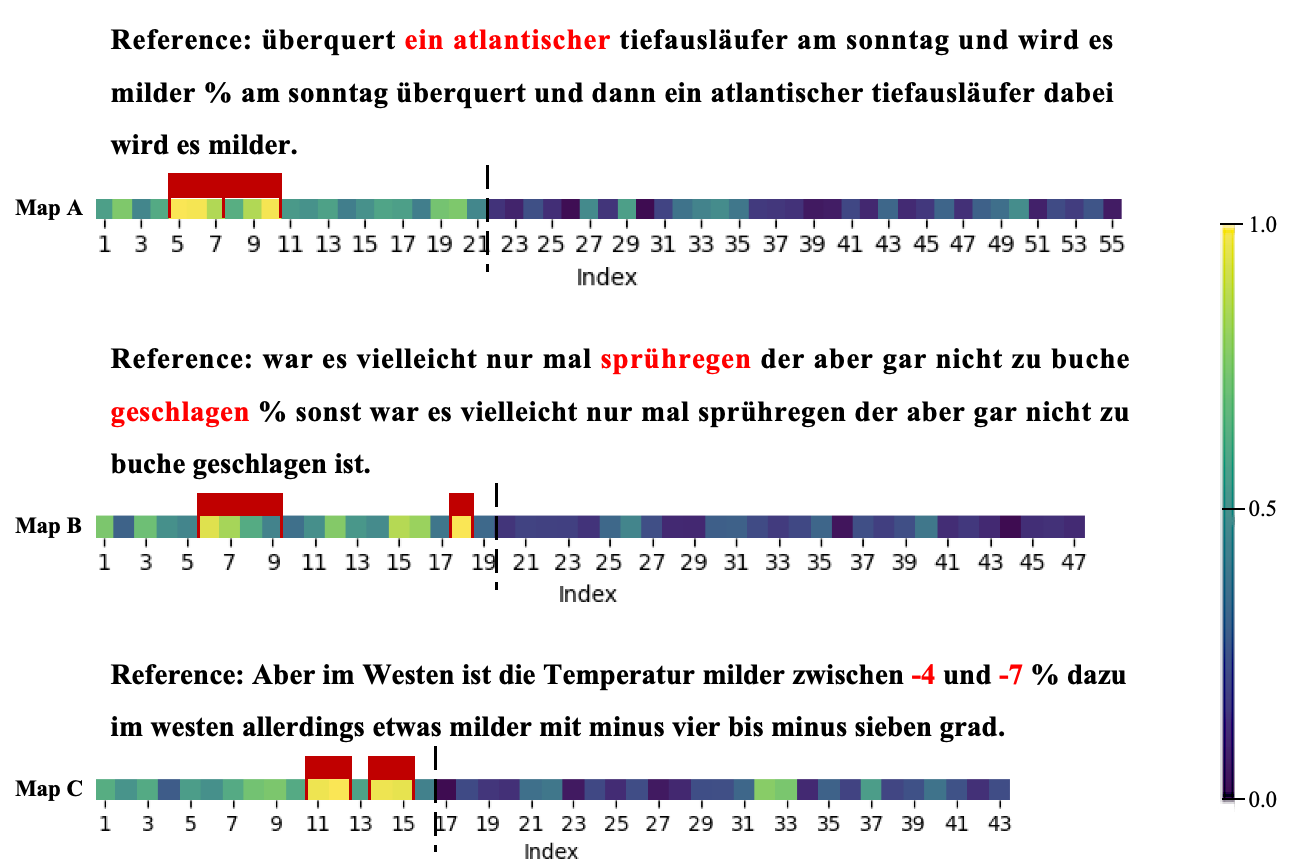}
	\caption{Visualization of weight heatmap, the brighter the color, the higher the weight.} 
	\label{fig:weights}
	\vspace{-10pt}
\end{figure}

\textbf{Visualization for SSAW.} 
To verify the weighted impact of the SSAW module on question text features, we visualized the weight map generated after processing the two sets of features using the SSAW module. As shown in Figure \ref{fig:weights}, the "Reference" consists of both the question text sequence and the ground-truth spoken text sequence, with the two sequences separated by a "\%". The "Map" represents the attention weight heatmap, where brighter colors indicate higher weights, and a black dashed line separates the question text sequence and the sign language video sequence.

The tokenizer divides the three question sequences into 21, 19, and 16 tokens, respectively. Considering the characteristics of German, a single word may be split into multiple tokens during tokenization. This means that multiple indices in Figure \ref{fig:weights} may jointly represent a single word. To clearly assess the importance of individual words, we compute the average weight of all tokens within a word and focus on the regions with higher attention weights (highlighted by red rectangles).

The SSAW module assigns higher weights to question features than video features, indicating that the question provides valuable supplementary information for SLT. When integrating question and sign language features, the model automatically identifies and selects key tokens, assigning them higher attention weights. In several examples in Figure \ref{fig:weights}, the model successfully recognized key tokens and assigned them higher weights. 
% This additional dialogue helps the model better understand the sign language video, allowing it to focus more on critical features, effectively narrow the translation scope, and ultimately improve translation quality.
These question cues enhance video comprehension by focusing on key features, narrowing translation possibilities to improve quality.

\textbf{Qualitative results.}
This paper presents a quantitative analysis of the PHOENIX-14T-QA dataset, providing an intuitive comparison of the impact of integrating question information and different integration methods on sign language translation performance. As shown in Table \ref{tab:q}, the "Reference" column represents the ground truth translations, against which the predictions of the three models are compared. Correct predictions are highlighted in green, while "[Miss]" indicates words contained in the "Reference" but missing in the prediction. Incorrect predictions are highlighted in red. If an incorrect or missing prediction occurs between two correct predictions, we consider the missing prediction to be preferable to an incorrect one.

By comparing the first example, we can observe that the translation results of QB-SLT contain fewer erroneous words than those of GFSLT-VLP, while SSL-SSAW produces translations with neither missing words nor incorrect predictions. From the second and third examples, we can observe that although SSL-SSAW may still encounter prediction errors and omissions, it remains the most effective approach among the three in terms of translation accuracy and semantic completeness.

Overall, integrating question context during translation contributes to constraining the model's translation space, thereby reducing the occurrence of erroneous words. Additionally, the proposed SSAW fusion module enhances the model's ability to leverage valuable auxiliary question information, further refining the translation space and minimizing errors. Furthermore, the introduced SSL strategy not only strengthens the model's generative capabilities but also improves semantic alignment between question and sign language, effectively reducing word omissions.

\begin{table}
	\caption{Quantitative analysis on the PHOENIX-2014T-QA dataset.}\label{tab:q}
	\resizebox{1.01\columnwidth}{!}{
		\renewcommand\arraystretch{1.3}
		\begin{tabular}{ l l}
			\Xhline{2pt}

			\textbf{Reference:}              & auch am wochenende schneit es zeitweise \\
			
			\textbf{SSL-SSAW:}  & \colorbox{green!30}{am wochenende schneit es zeitweise} \\
			
			\textbf{DB-GFSLT:} & \colorbox{red!30}{schneit es}  \colorbox{green!30}{auch am wochenende} \colorbox{yellow!35}{\textbf{[Miss]} \textbf{[Miss]}} \colorbox{green!30}{zeitweise}\\
			
			\textbf{GFSLT-VLP:} & \colorbox{yellow!35}{\textbf{[Miss]}} \colorbox{green!30}{am wochenende} \colorbox{red!30}{fällt hier und da etwas regen oder schnee}\\

			\hline
			
			\textbf{Reference:}              & heute nacht gibt es an der oder noch einzelne schneeschauer \\
			
			\textbf{SSL-SSAW:}         & \colorbox{green!30}{heute nacht gibt es} \colorbox{red!30}{verbreitet} \colorbox{yellow!35}{\textbf{[Miss]} \textbf{[Miss]} \textbf{[Miss]} \textbf{[Miss]}} \colorbox{green!30}{schneeschauer}\\
			
			\textbf{DB-GFSLT:} & \colorbox{yellow!35}{\textbf{[Miss]} \textbf{[Miss]}} \colorbox{green!30}{gibt es}  \colorbox{red!30}{heute abend eine} \colorbox{yellow!35}{\textbf{[Miss]} \textbf{[Miss]}} \colorbox{green!30}{schneeschauer}  \colorbox{red!30}{gibt}\\
			
			\textbf{GFSLT-VLP:} &  \colorbox{green!30}{heute nacht}  \colorbox{red!30}{schneit} \colorbox{green!30}{es}  \colorbox{red!30}{stellenweise im norden auch schneeregen} \colorbox{yellow!35}{\textbf{[Miss]}}\\
			
			\hline

			\textbf{Reference:}              & am tag ist es meist stark bewölkt oder neblig trüb \\
			
			\textbf{SSL-SSAW:}         & \colorbox{green!30}{am tag} \colorbox{yellow!35}{\textbf{[Miss]} \textbf{[Miss]}} \colorbox{green!30}{meist stark bewölkt oder neblig trüb}\\
			
			\textbf{DB-GFSLT:} & \colorbox{red!30}{ist es} \colorbox{green!30}{am tag} \colorbox{yellow!35}{\textbf{[Miss]} \textbf{[Miss]}} \colorbox{green!30}{meist stark bewölkt oder neblig trüb} \\ &  \colorbox{red!30}{am alpenrand regnet oder neblig}\\
			
			\textbf{GFSLT-VLP:} &  \colorbox{green!30}{am tag}  \colorbox{red!30}{gibt}  \colorbox{green!30}{es} \colorbox{red!30}{neben wolken und nebelfeldern auch gebietsweise wolken}\\

			\Xhline{2pt}

		\end{tabular}
	}
	\vspace{-15pt}
\end{table}

\section{Conclusion}
In the real world, communication between deaf people and hearing people is often accompanied by the generation of question information. This paper introduces the integration of such question information into sign language translation for the first time, replacing traditional annotations (such as glosses) that are difficult to obtain, thereby enhancing the translation effectiveness. Furthermore, by leveraging cross-modality feature fusion and effectively utilizing known question information for training, this study improves the model's contextual understanding and generalization capabilities.

\section*{Declaration of Interests Statement}
The authors declare that they have no known competing financial interests or personal relationships that could have appeared to influence the work reported in this paper.

\section*{Acknowledgment}
The work is supported by the National Defense Science and Technology Innovation Fund of the Chinese Academy of Sciences (No. 19-H863-00-KX-001-002-01), and the National Natural Science Foundation of China (Grant No. 62072334).

\bibliographystyle{elsarticle-num-names} 
\bibliography{ref}
\end{document}